\documentclass[letterpaper]{article} 
\usepackage{aaai2026}  
\usepackage{times}  
\usepackage{helvet}  
\usepackage{courier}  
\usepackage[hyphens]{url}  
\usepackage{graphicx} 
\usepackage{natbib}  
\usepackage{caption} 
\usepackage{algorithm}
\usepackage{algorithmic}

\usepackage{newfloat}
\usepackage{listings}
\DeclareCaptionStyle{ruled}{labelfont=normalfont,labelsep=colon,strut=off} 
\floatstyle{ruled}
\newfloat{listing}{tb}{lst}{}
\floatname{listing}{Listing}
\usepackage[colorlinks=true]{hyperref} 

\usepackage{xcolor}
\usepackage{amsmath}
\usepackage{makecell,pifont,multirow,multicol}
\usepackage{colortbl}
\usepackage{caption}
\usepackage{subcaption}
\usepackage{booktabs}
\usepackage{amssymb}
\usepackage{marvosym}
\usepackage{titletoc}

\def\netName{Polaris}

\title{Relative Position Matters: Trajectory Prediction and Planning with Polar Representation}
\author {
    Bozhou Zhang\textsuperscript{\rm 1}\quad
    Nan Song\textsuperscript{\rm 1}\quad
    Bingzhao Gao\textsuperscript{\rm 2}\quad
    Li Zhang\textsuperscript{\rm 1}\corrauthor
}
\affiliations {
    \textsuperscript{\rm 1}School of Data Science, Fudan University\quad
    \textsuperscript{\rm 2}Tongji University\\
    \vspace{.4em} 
}

\begin{document}

\maketitle


\begin{abstract}

Trajectory prediction and planning in autonomous driving are highly challenging due to the complexity of predicting surrounding agents' movements and planning the ego agent's actions in dynamic environments. 
Existing methods encode map and agent positions and decode future trajectories in Cartesian coordinates. However, modeling the relationships between the ego vehicle and surrounding traffic elements in Cartesian space can be suboptimal, as it does not naturally capture the varying influence of different elements based on their relative distances and directions.
To address this limitation, we adopt the Polar coordinate system, where positions are represented by radius and angle. This representation provides a more intuitive and effective way to \textit{model spatial changes and relative relationships}, especially in terms of distance and directional influence.
Based on this insight, we propose \textbf{\netName{}}, a novel method that operates entirely in Polar coordinates, distinguishing itself from conventional Cartesian-based approaches.
By leveraging the Polar representation, this method explicitly models distance and direction variations and captures relative relationships through dedicated encoding and refinement modules, enabling more structured and spatially aware trajectory prediction and planning.
Extensive experiments on the challenging prediction (Argoverse 2) and planning benchmarks (nuPlan) demonstrate that \netName{} achieves state-of-the-art performance.

\end{abstract}

\section{Introduction}
\label{sec:intro}


\begin{figure}[t!]
\centering
\includegraphics[width=0.47\textwidth]{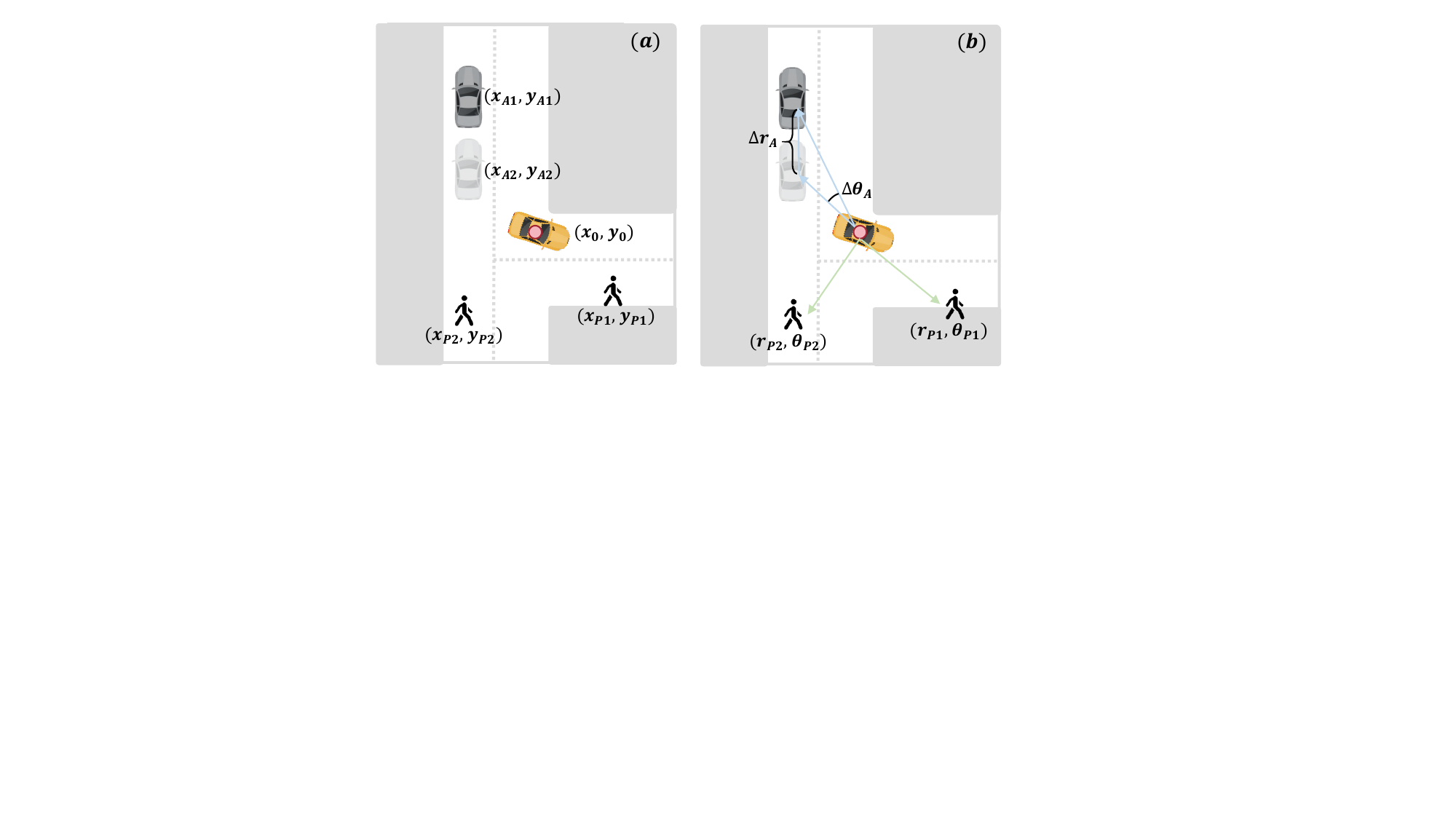}
\caption{
\textbf{Comparison between our \netName{} and previous methods.}
As illustrated in \textbf{(a)}, previous methods encode agents and lanes to predict future trajectories in Cartesian coordinates. In contrast, our \netName{}, shown in \textbf{(b)}, operates entirely in the polar coordinate system, \textit{explicitly representing radius and angle to intuitively model relative distances, directions, and their variations}. This leads to more accurate trajectory prediction and planning.
}
\label{fig:first}
\end{figure}

Trajectory prediction and planning, which involve predicting the future trajectories of surrounding vehicles and pedestrians and planning the ego agent's future trajectory using HD maps and historical trajectory data, are crucial for autonomous driving~\cite{argoverse2,nuscenes}. 
In recent years, significant progress has been made in this area. Some studies focus on improving the encoding of map and historical trajectory data to enhance scene context features~\cite{vectornet,SceneTransformer}, while others explore better methods for decoding future trajectories~\cite{mtr,qcnet,DeMo,plantf}.

Existing methods~\cite{mtr,mtr++,qcnet,plantf} follow a paradigm based on the Cartesian coordinate system. In these approaches, the scene context is encoded in Cartesian coordinates, and future trajectory positions are regressed accordingly, as illustrated in Figure~\ref{fig:first}~(a).
These methods use $(x, y)$ positions to model the relationships between the ego vehicle and surrounding traffic elements. The varying importance of these elements—determined by their relative distances and directions to the ego vehicle—is learned implicitly. Some approaches attempt to capture such relationships using $(\Delta x, \Delta y)$; however, this form still does not directly encode relative distance and direction, instead relying on the model to infer them. Such implicit modeling may lead to suboptimal performance.
For example, as shown in Figure~\ref{fig:first}, pedestrian 1 (on the right) is directly ahead of the vehicle, while pedestrian 2 (on the left) is positioned to the side. Pedestrian 1 should have a greater influence on the ego vehicle’s motion estimation than pedestrian 2. In Cartesian-based methods, such differences are not explicitly modeled but instead must be inferred through learning.
In contrast, the Polar coordinate system provides an explicit representation of relative distances and directions. This enables the model to directly capture the varying influence of traffic elements in a more structured manner, leading to more accurate motion estimation.
This observation motivates the design of new approaches that can better encode the variations and relative relationships among traffic elements by leveraging the advantages of Polar representation.

Motivated by the above insights, we propose a novel framework, \textbf{\netName{}}, for trajectory prediction and planning based on Polar representation.
As shown in Figure~\ref{fig:first}~(b), both the input and output are represented in the Polar coordinate system. By leveraging the intuitive nature of Polar coordinates—expressed as $(r, \theta)$—our approach effectively models the relative distances and directions between the target agent and surrounding traffic elements, as well as their temporal variations.
The proposed framework consists of two main components: Polar scene context encoding and Polar relationship refinement.
The Polar scene context encoding module encodes the motion states (e.g., position, velocity, acceleration) of agents, as well as lane geometry and lane-change information from HD maps. It captures the relative relationships and spatial dynamics within the scene context.
The Polar relationship refinement module refines the proposal output trajectories by interacting with surrounding agents and the map, further modeling the relative relationships within the scene. This refinement enhances the accuracy of prediction and planning.
To support this process, we design a Relative Embedding Transformer, which explicitly embeds relative distances and angles, enabling more effective relationship modeling within the Polar representation.

Our \textbf{contributions} are summarized as follows:
{\bf (i)} We are the first to perform trajectory prediction and planning entirely in the Polar coordinate.
{\bf (ii)} We propose \netName{}, a Polar-based framework that incorporates Polar scene context encoding and Polar relationship refinement, both equipped with Relative Embedding Transformer to model relative spatial relationships.
{\bf (iii)} Extensive experiments on the Argoverse 2 and nuPlan benchmarks demonstrate that \netName{} achieves state-of-the-art performance.

\section{Related work}

\paragraph{Trajectory prediction.}
Recent advancements in autonomous driving highlight the importance of accurately predicting agent behavior through effective scene representation. Early methods~\cite{multipath,covernet,home} typically converted driving scenarios into image-based formats and applied convolutional networks for context encoding, but often struggled to capture fine-grained structural details. This limitation motivated a transition to vectorized representations~\cite{vectornet,multipath++,hivt,densetnt}, enabling more structured and geometry-aware modeling of the environment.
Building upon these representations, various frameworks have been proposed to predict multi-modal trajectories. Initial approaches focused on goal-oriented strategies~\cite{densetnt,DSP} and probability heatmaps~\cite{home,gohome}, while more recent models such as MTR~\cite{mtr}, QCNet~\cite{qcnet}, and others~\cite{wayformer,SceneTransformer,most,HPTR,frm,modeseq,eda,retromotion} leverage Transformer architectures~\cite{attention} to better capture scene-agent interactions. These advances are further enhanced by emerging paradigms, including graph-based modeling~\cite{hdgt,LaneGCN,fjmp}, pre-training~\cite{Forecast-mae,sept,smartpretrain}, history-aware designs~\cite{t4p,hpnet}, GPT-style decoding~\cite{trajeglish,motionlm,donut,drivegpt}, reinforcement learning~\cite{foresightmotion}, flow matching~\cite{flowmatch}, and post-refinement techniques~\cite{R-Pred,smartrefine}.
In parallel, there has been increasing attention to multi-agent prediction, where models predict trajectories for all agents jointly, as explored in~\cite{mtr++,SceneTransformer,smart,catk,thomas,optimizing}, aiming to enhance consistency and interaction reasoning in complex driving scenarios.
Some other works~\cite{unitraj} aim to unify benchmarks~\cite{waymo,nuplan,argoverse,argoverse2,nuscenes} by establishing a standardized training and evaluation protocol.

\paragraph{Trajectory planning.}
As the final stage of autonomous driving, trajectory planning plays a key role in ensuring safe and precise navigation.
A classic rule-based method is the Intelligent Driver Model (IDM)~\cite{IDM}, which follows a leading vehicle while maintaining a safe distance.
The release of the nuPlan~\cite{nuplan} dataset and its standardized simulation benchmark has paved the way for advancing learning-based motion planners.
Recent works explore purely learning-based approaches~\cite{plantf,pluto,betop,diffusionplanner,int2planner}, while others combine learning and rule-based methods for hybrid planning~\cite{PDM}. 
Some methods explore accurate planning by leveraging large language models~\cite{planagent,llmplan,instructplan,zheng2025driveagent,padriver,calmm,asynchronous}, world models~\cite{planwithWM,xiao2024learning}, tree policy~\cite{dtpp}, generative models~\cite{gump,gpd-1}, diffusion models~\cite{diffusionplanner,diffusiones}, mixture of experts~\cite{moeplan}, POMDP planning~\cite{pomdp}, and reinforcement learning~\cite{carplanner,planr1,gendrive,carl}.


\begin{figure*}[t!]
\centering
\includegraphics[width=1\textwidth]{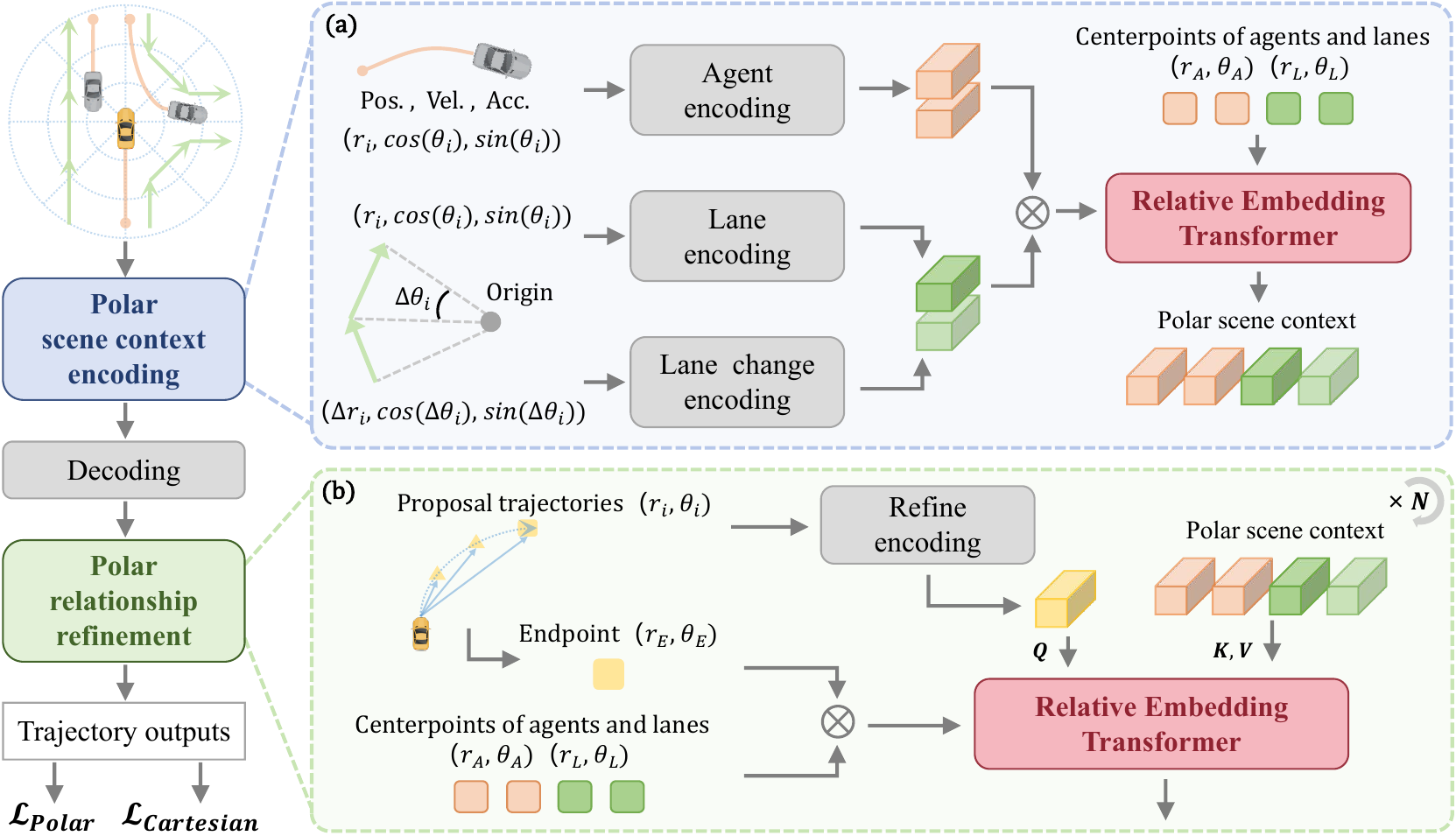}
\caption{
Overview of our \textbf{\netName{}} framework. (a) The Polar scene context encoding separately encodes the motion states of agents, lanes, and lane changes, all within the Polar coordinate. It then employs the Relative Embedding Transformer to model interactions and derive the scene context. Subsequently, the decoding module predicts the proposal future trajectories in Polar representation. (b) The Polar relationship refinement then re-encodes the proposal trajectories and interacts with the scene context using the Relative Embedding Transformer to produce the final output. Losses are computed in both Cartesian and Polar coordinate.
}
\label{fig:main}
\end{figure*}

\paragraph{Polar representation in autonomous driving.} 
While Cartesian representation is commonly used, Polar representation is favored in some methods for its inherent advantages in autonomous driving. In LiDAR-based object detection~\cite{polarnet,polarstream,partner}, Polar coordinates naturally align with the radial distribution of LiDAR data and the distance-dependent density variation in point clouds, allowing for more effective modeling of these characteristics.
PolarFormer~\cite{polarformer} further leverages a cross-attention mechanism to capture the geometric structure of BEV features in Polar space, achieving strong performance in camera-based object detection.
However, the potential of Polar representation in trajectory prediction and planning remains underexplored. To the best of our knowledge, we are the first to introduce it in this context, and our method achieves notable improvements.

\section{Methodology}

\begin{figure}[t!]
\centering
\includegraphics[width=0.47\textwidth]{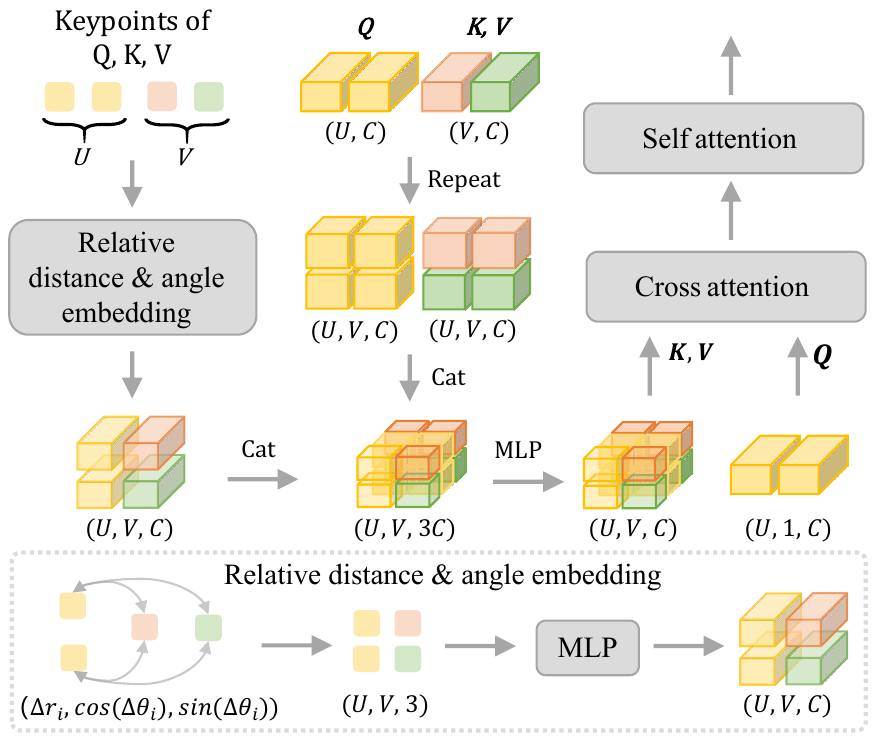}
\caption{
Overview of our \textbf{Relative Embedding Transformer}. We use $U$ queries and $V$ keys and values as an example to illustrate the feature dimensions. For simplicity, the batch size dimension is omitted.
}
\label{fig:trans}
\end{figure}

In this section, we introduce \textbf{\netName{}}, a novel framework for trajectory prediction and planning using Polar representation, as illustrated in Figure~\ref{fig:main}. The architecture comprises key components, including Polar scene context encoding and Polar relationship refinement, both of which leverage a Relative Embedding Transformer, as shown in Figure~\ref{fig:trans}. It also incorporates loss calculation in both Cartesian and Polar coordinate systems.

\subsection{Problem formulation}
Trajectory prediction and planning involve predicting the future trajectories of surrounding agents and planning the ego agent’s path, based on the HD map and historical agent trajectories. We employ a vectorized representation, similar to that used in other methods~\cite{vectornet,mtr}. For the input, the HD map consists of several lane instances, each lane is composed of multiple points. Specifically, this can be represented as ${M} \in \mathbb{R}^{N_{\rm m} \times L \times C_{\rm m}}$, where $N_{\rm m}$, $L$, and $C_{\rm m}$ denote the number of lane instances, points per lane, and feature channels (e.g., position), respectively. Agents, which include traffic participants such as vehicles and pedestrians, are represented as ${A} \in \mathbb{R}^{N_{\rm a} \times T_{\rm h} \times C_{\rm a}}$. Here, $N_{\rm a}$, $T_{\rm h}$, and $C_{\rm a}$ stand for the number of agents, historical timestamps, and motion states (e.g., position, velocity, acceleration). For the output, the model predicts the future trajectories ${A_{\rm f}} \in \mathbb{R}^{K \times N_{\rm aoi} \times T_{\rm f} \times 2}$ for the agents of interest, where $K$, $N_{\rm aoi}$, $T_{\rm f}$ indicate the modalities, the number of agents of interest, and future timestamps, respectively. The associated probabilities ${P_{\rm f}} \in \mathbb{R}^{K \times N_{\rm aoi}}$ are predicted as well. We utilize Polar coordinates to represent position, velocity, acceleration, and other attributes as $(r, \theta)$ rather than $(x, y)$.


\subsection{Polar scene context encoding}
As shown in Figure~\ref{fig:main} (a), when we obtain the Polar representation of the HD map as $M$ and agents as $A$, we need to encode them to effectively capture the scene context. To enhance the model's ability to learn, we input the Polar coordinates in the form of $(r, cos\theta, sin\theta)$. For the HD map, we utilize a PointNet-based lane encoder, as described in previous works~\cite{mtr, Forecast-mae}. To leverage the advantages of Polar representation for modeling relative relationships, we compute the difference between the coordinates of adjacent lane points to obtain the lane change variable, denoted as $dM \in \mathbb{R}^{N_{\rm m} \times (L-1) \times C_{\rm m}}$. We use separate encoders to extract the map features $F_{\rm m} \in \mathbb{R}^{N_{\rm m} \times C}$ and $F_{\rm dm} \in \mathbb{R}^{N_{\rm m} \times C}$; we then combine them to derive the final map feature $F_{\rm m}$. The process can be formulated as:

\begin{equation}
\begin{split}
F_{\rm m} &= {\rm PointNet_M}({M}),\\
F_{\rm dm} &= {\rm PointNet_{dM}}({dM}), \\
F_{\rm m} &= {\rm MLP}({\rm Concat}(F_{\rm m},\, F_{\rm dm})).
\end{split}
\end{equation}

For the agents, we aggregate the historical trajectory features up to the current time step, and utilize Mamba~\cite{mamba} blocks to extract features $F_{\rm a} \in \mathbb{R}^{N_{\rm a} \times C}$, given their exceptional capability for efficient and effective sequence modeling, as described in the previous work~\cite{DeMo}. Subsequently, the scene context features $F_{\rm s} \in \mathbb{R}^{(N_{\rm a} + N_{\rm m}) \times C}$ are formed by combining the agent and map features, which are then propagated through the Relative Embedding Transformer for intra-interaction learning. The centerpoints of lane instances and agents (i.e., the middle point of each lane instance and the current position of each agent) are also input into the Relative Embedding Transformer. Further details are discussed below in the Relative Embedding Transformer section. The overall process can be formulated as:

\begin{equation}
\begin{split}
F_{\rm a} &= {\rm Mamba}({A}),\\
F_{\rm s} &= {\rm Concat}(F_{\rm m},\, F_{\rm a}), \\
F_{\rm s} &= {\rm Transformer}(F_{\rm s}).
\end{split}
\end{equation}

\subsection{Trajectory decoding}
After obtaining the scene context features, we aim to decode multi-modal future trajectories for the agents of interest. We utilize vanilla Transformer blocks to facilitate interactions between the multi-modal trajectory queries and the scene context features. Subsequently, an MLP is employed to decode the proposal trajectories ${A_{\rm f_{prop}}} \in \mathbb{R}^{K \times N_{\rm aoi} \times T_{\rm f} \times 2}$ in the Polar coordinate system, represented by $(r, \theta)$, along with their associated probabilities ${P_{\rm f_{prop}}} \in \mathbb{R}^{K \times N_{\rm aoi}}$.

\subsection{Polar relationship refinement}
As shown in Figure~\ref{fig:main} (b), to further enhance the interaction between future trajectories and scene context features, we design Polar relationship refinement modules to refine the proposal trajectories. The proposal trajectories are re-encoded with an MLP to obtain the multi-modal trajectory queries $Q_{\rm traj} \in \mathbb{R}^{K \times N_{\rm aoi} \times C}$. Additionally, the endpoints $P_{\rm E} (r_{\rm E}, \theta_{\rm E})$ of the multi-modal trajectories, which are crucial for the accuracy of the overall trajectory, are extracted and combined with the centerpoints of lane instances and agents. The centerpoints are the same as those discussed in the encoder section. These keypoints of future trajectories and scene context features are used to model relative relationships within the Relative Embedding Transformer. Then, the trajectory queries $Q_{\rm traj}$, the scene context features $F_{\rm s}$, and the keypoints of query, key and value are input into the Relative Embedding Transformer. Finally, an MLP is used to obtain the refined outputs, including trajectories ${A_{\rm f_{ref}}}$ and probabilities ${P_{\rm f_{ref}}}$. The refinement module is iterated multiple times to achieve the final results. The refinement process can be described as:

\begin{equation}
\begin{split}
Q_{\rm traj} &= {\rm MLP}({A_{\rm f_{prop}}}),\\
Q_{\rm traj} &= {\rm Transformer}({\rm Q} = Q_{\rm traj}, {\rm K,V} = F_{\rm s}),\\
A_{\rm f_{ref}}, P_{\rm f_{ref}} &= {\rm MLP}(Q_{\rm traj}).
\end{split}
\end{equation}


\begin{table*} [t!]

\centering
\setlength{\tabcolsep}{2.05mm}
{\begin{tabular}{l|cccccc}
\toprule[1.5pt]
Method & $\textit{minFDE}_{1}$ $\downarrow$ & $\textit{minADE}_{1}$ $\downarrow$ & $\textit{minFDE}_{6}$ $\downarrow$ & $\textit{minADE}_{6}$ $\downarrow$ & $\textit{MR}_{6}$ $\downarrow$ & $\textit{b-minFDE}_{6}$ $\downarrow$\\
[1.5pt]\hline\noalign{\vskip 2pt}
GoRela~\cite{gorela}            & 4.62 & 1.82 & 1.48 & 0.76 & 0.22 & 2.01\\ 
MTR\cite{mtr}                   & 4.39 & 1.74 & 1.44 & 0.73 & \underline{0.15} & 1.98\\
HPTR~\cite{HPTR}                & 4.61 & 1.84 & 1.43 & 0.73 & 0.19 & 2.03\\
SIMPL~\cite{simpl}              & 5.50 & 2.03 & 1.43 & 0.72 & 0.19 & 2.05\\
GANet~\cite{ganet}              & 4.48 & 1.77 & 1.34 & 0.72 & 0.17 & 1.96\\
ProphNet~\cite{prophnet}        & 4.74 & 1.80 & 1.33 & 0.68 & 0.18 & 1.88\\  
QCNet~\cite{qcnet}              & 4.30 & 1.69 & 1.29 & 0.65 & 0.16 & 1.91\\  
SmartRefine~\cite{smartrefine}  & 4.17 & 1.65 & \underline{1.23} & \underline{0.63} & \underline{0.15} & \underline{1.86}\\
RealMotion~\cite{RealMotion}    & \underline{3.93} & \underline{1.59} & 1.24 & 0.66 & \underline{0.15} & 1.89 \\
\rowcolor{gray!20}
\bf\netName~(Ours) & \bf 3.78 &  \bf 1.53 & \bf 1.15 & \bf 0.62 &\bf 0.13 & \bf 1.80 \\
[1.5pt] \hline\noalign{\vskip 2pt}

MacFormer~\cite{macformer}       & 4.69 & 1.84 & 1.38 & 0.70 & 0.19 & 1.90\\
Gnet~\cite{gnet}                 & 4.40 & 1.72 & 1.34 & 0.69 & 0.18 & 1.90 \\
QCNet~\cite{qcnet}               & 3.96 & 1.56 & \underline{1.19} & 0.62 & \underline{0.14} & 1.78\\  
DyMap~\cite{DyMap}               & 3.99 & 1.59 & 1.21 & 0.66 & 0.15 & 1.78 \\
DeMo~\cite{DeMo}               & \bf 3.70 & \bf 1.49 & \bf 1.11 & \bf 0.60 & \bf 0.12 & \underline{1.73} \\
\rowcolor{gray!20}
\bf\netName~(Ours)               & \underline{3.78} & \underline{1.53} & \bf 1.11 & \underline{0.61} & \bf 0.12 & \bf 1.71 \\
\bottomrule[1.5pt]
\end{tabular}}
\caption{
Performance of trajectory prediction \textit{on the Argoverse 2 single-agent test set from the official leaderboard}. For each metric, the best result is highlighted in \textbf{bold}, and the second-best is \underline{underlined}. The upper section reports results from single-model, while the lower section includes results with model ensembling.
}
\label{tab:av2}
\end{table*}


\begin{table*} [ht!]

\centering
\setlength{\tabcolsep}{0.95mm}
{\begin{tabular}{l|ccccc}
\toprule[1.5pt]
Method & $\textit{avgMinFDE}_{1}$ $\downarrow$ & $\textit{avgMinADE}_{1}$ $\downarrow$ & $\textit{avgMinFDE}_{6}$ $\downarrow$ & $\textit{avgMinADE}_{6}$ $\downarrow$ & $\textit{actorMR}_{6}$ $\downarrow$ \\[1.5pt]\hline\noalign{\vskip 2pt}
FJMP~\cite{fjmp} & 4.00 & 1.52 & 1.89 & 0.81 & 0.23 \\
Forecast-MAE~\cite{Forecast-mae} & 3.33 & 1.30 & 1.55 & 0.69 & 0.19 \\
RealMotion~\cite{RealMotion} & 2.87 & 1.14 & 1.32 & 0.62 & 0.18 \\
DeMo~\cite{DeMo} & \underline{2.78} & \underline{1.12} & \underline{1.24} & \underline{0.58} & \underline{0.16} \\[1.5pt]
\rowcolor{gray!20}
\bf\netName~(Ours) & \bf 2.67 & \bf 1.07 & \bf 1.18 & \bf 0.56 & \bf 0.15 \\
\bottomrule[1.5pt]
\end{tabular}}
\caption{
Performance of trajectory prediction \textit{on the Argoverse 2 multi-agent test set from the official leaderboard}.
}
\label{tab:av2_multi}

\end{table*}

\subsection{Relative Embedding Transformer}
As shown in Figure~\ref{fig:trans}, taking full advantage of the strengths of Polar representation, the Relative Embedding Transformer is designed to embed the relative positions of the query, key, and value during interactions. First, the relative distances and angles of the keypoints (e.g., centerpoints for lane instances and agents, endpoints for future trajectories) in the query, key, and value are calculated in the form of $(\Delta r, cos(\Delta\theta), sin(\Delta\theta))$. An MLP is then used to encode these relative positions into relative position features. The relative relationships among the query, key, and value are determined by relative distances and angles of their keypoints in this manner. The query, key, and value features are repeated to match the size of the relative position features. The query, key, and value features, along with the relative position features, are then concatenated and passed through an MLP to adjust their dimensions, forming the new key and value. Finally, cross-attention and self-attention mechanisms are applied to obtain the output features. In the encoder part, the query, key, and value all correspond to the scene context features $F_{\rm s}$. In the refinement part, the query consists of the multi-modal trajectory queries $Q_{\rm traj}$, while the key and value remain the scene context features $F_{\rm s}$.

\subsection{Training losses}
During the training process, predicted trajectories are supervised using the smooth-l1 loss $\mathcal{L}_{\rm reg}$ for regression, while the associated probabilities are supervised using the cross-entropy loss $\mathcal{L}_{\rm cls}$ for classification. Losses are calculated for both the proposal outputs and the refinement outputs. The total loss comprises two parts: the Polar loss, derived from the direct outputs in Polar coordinates $(r, \theta)$, and the Cartesian loss, derived from transforming the outputs into Cartesian coordinates $(x, y)$. All losses are weighted equally, and the winner-take-all strategy is employed, optimizing only the best prediction with the minimal average prediction error compared to the ground truth. The overall process can be formulated as follows:

\begin{equation}
\begin{split}
\mathcal{L_{\rm proposal/refine}} &= \mathcal{L}_{\rm reg} + \mathcal{L}_{\rm cls}, \\
\mathcal{L_{\rm Polar/Cartesian}} &= \mathcal{L}_{\rm proposal} + \mathcal{L}_{\rm refine}, \\
\mathcal{L_{\rm total}} &= \mathcal{L}_{\rm Polar} + \mathcal{L}_{\rm Cartesian}. \\
\end{split}
\end{equation}

\section{Experiments}

\subsection{Experimental settings}

\paragraph{Datasets and evaluation metrics.} 
We evaluate our method on two widely used and challenging datasets: the Argoverse 2~\cite{argoverse2} dataset for trajectory prediction and the nuPlan~\cite{nuplan} dataset for trajectory planning.
The Argoverse 2 dataset is sampled at 10 Hz and provides 5 seconds of historical trajectories along with 6 seconds of future predictions.
The nuPlan dataset uses a simulator that runs each scenario for 15 seconds at 10 Hz.
For trajectory prediction, we use standard metrics, including $minADE$, $minFDE$, $MR$, and $b\text{-}minFDE$.
For trajectory planning, nuPlan evaluates performance using three key metrics: the open-loop score (OLS), the non-reactive closed-loop score (NR-CLS), and the reactive closed-loop score (R-CLS).
The evaluations cover 6 modes for both Argoverse 2 and nuPlan. Further details are provided in the appendix.


\begin{table*} [t!]

\centering
{
\begin{tabular}{l|l|ccc}
\toprule[1.5pt]
Type & Method & OLS $\uparrow$ & NR-CLS $\uparrow$ & R-CLS $\uparrow$ \\
[1.5pt]\hline\noalign{\vskip 2pt}

\multirow{2}{*}{Rule} & IDM~\cite{IDM} & 0.20 & 0.56 & 0.62 \\
                      & PDM-Closed~\cite{PDM} & 0.26 & 0.65 & 0.75 \\
\midrule
\multirow{2}{*}{Hybrid} & GameFormer~\cite{gameformer} & 0.75 & 0.67 & 0.69 \\
                        & PDM-Hybrid~\cite{PDM} & 0.74 & 0.66 & 0.76 \\
\midrule
\multirow{7}{*}{Learning} & UrbanDriver~\cite{UrbanDriver} & 0.77 & 0.52 & 0.49 \\
                          & PDM-Open~\cite{PDM} & 0.79 & 0.34 & 0.36 \\   
                          & PlanCNN~\cite{plant} & 0.52 & 0.49 & 0.52 \\
                          & GC-PGP~\cite{GCPGP} & 0.74 & 0.43 & 0.40 \\
                          & PlanTF~\cite{plantf} & 0.83 & 0.73 & 0.62 \\
                          & BeTopNet~\cite{betop} & \underline{0.84} & \bf 0.77 & \underline{0.69} \\
                          \rowcolor{gray!20}
                          & \bf\netName~(Ours) & \bf 0.86 & \underline{0.74} & \bf 0.70 \\
\bottomrule[1.5pt]
\end{tabular}
}
\caption{
Performance of open-loop and closed-loop planning \textit{on the nuPlan dataset under the Test 14 Hard benchmark}.
}
\label{tab:nuplan}

\end{table*}


\begin{table*} [t!]

\centering
{\begin{tabular}{c|ccc|ccccc}

\toprule[1.5pt]
ID & A.L.C. Embed & Polar Refine & Rel. Transf. & $\textit{minFDE}_{1}$ & $\textit{minADE}_{1}$ &  $\textit{minFDE}_{6}$ & $\textit{minADE}_{6}$ & $\textit{MR}_{6}$\\
[1.5pt]\hline\noalign{\vskip 2pt}

1 &  &  &  & 4.51 & 1.79 & 1.48 & 0.76 & 0.21\\ 
2 & \checkmark &  &  & 4.46 & 1.78 & 1.40 & 0.72 & 0.19\\
3 & \checkmark  & \checkmark &  & 4.26 & 1.69 & 1.35 & 0.70 & 0.18\\

\rowcolor{gray!20}
4 & \checkmark & \checkmark & \checkmark & 3.88 & 1.55 & 1.21 & 0.63 & 0.14 \\
\bottomrule[1.5pt]
\end{tabular}}
\caption{Ablation study on the core components of \netName{} \textit{on the Argoverse 2 validation set}. ``A.L.C. Embed" indicates agent and lane change embedding. ``Polar Refine" indicates Polar relationship refinement module. ``Rel. Transf." indicates Relative Embedding Transformer.}
\label{tab:abl_main}

\end{table*}

\begin{table} [t!]

\centering
{\begin{tabular}{cc|ccc}
\toprule[1.5pt]
Cartesian & Polar & \multirow{2}{*}{$\textit{minFDE}_{6}$} & \multirow{2}{*}{$\textit{minADE}_{6}$} & \multirow{2}{*}{$\textit{MR}_{6}$}\\
Loss & Loss & & & \\
[1.5pt]\hline\noalign{\vskip 2pt}

\checkmark &  & 1.29 & 0.68 & 0.16\\
& \checkmark & 1.34 & 0.71 & 0.17\\

\rowcolor{gray!20}
\checkmark & \checkmark &  1.21 &  0.63 &  0.14 \\
\bottomrule[1.5pt]
\end{tabular}}
\caption{Ablation study on the losses.}
\label{tab:abl_loss}

\end{table}

\paragraph{Implementation details.} 
We train our models for 80 epochs using the AdamW optimizer, with a batch size of 4 per GPU. The training process is end-to-end with a learning rate of 0.001 and a weight decay of 0.01. Additionally, we use a cosine learning rate schedule with a 10-epoch warm-up phase. 
Experiments are conducted on 8 NVIDIA GeForce RTX 3090 GPUs. 
As for the number of layers in each component, the encoding module has 3 Relative Embedding Transformer layers, the decoding module has 2 vanilla Transformer layers, and there are 2 refining modules, each with 2 Relative Embedding Transformer layers. 
Additional configuration details and further experiments are provided in the appendix.

\subsection{Comparison with state of the art}
We compare \netName{} with several recent models on the Argoverse 2 dataset, as shown in Table~\ref{tab:av2}. To ensure a fair comparison, the performance is evaluated both with and without model ensembling. In the single-model setting, the results show that \netName{} outperforms all previous methods across all metrics, including state-of-the-art models such as QCNet~\cite{qcnet}, SmartRefine~\cite{smartrefine}, and RealMotion~\cite{RealMotion}.
When employing ensembling techniques similar to those used by other methods, \netName{} further improves its performance, significantly surpassing most existing approaches across all metrics and achieving results comparable to DeMo~\cite{DeMo}.
We also evaluate \netName{} under the multi-agent setting, as reported in Table~\ref{tab:av2_multi}, where it continues to demonstrate strong performance.

For the trajectory planning task, we compare \netName{} with other top-performing methods on the nuPlan dataset under the Test 14 Hard benchmark, as shown in Table~\ref{tab:nuplan}. As a purely learning-based method with no hand-crafted rules, \netName{} outperforms state-of-the-art models such as PlanTF~\cite{plantf} and BeTopNet~\cite{betop}, and achieves competitive performance compared to the rule-based method PDM-Closed~\cite{PDM}.


\begin{table} [t!]

\centering
{\begin{tabular}{c|cccc}
\toprule[1.5pt]
Coordinate & $\textit{minFDE}_{6}$ & $\textit{minADE}_{6}$ & $\textit{MR}_{6}$ & Inf.\\
[1.5pt]\hline\noalign{\vskip 2pt}

Cartesian$_{ori}$ & 1.30 & 0.68 & 0.16 & 110ms\\
Cartesian$_{mod}$ & 1.33 & 0.69 & 0.16 & 67ms\\

\rowcolor{gray!20}
Polar & 1.21 & 0.63 & 0.14 & 48ms\\

\bottomrule[1.5pt]
\end{tabular}}
\caption{Ablation study on the coordinate systems. ``Inf." indicates inference speed.}
\label{tab:carpol}

\end{table}

\subsection{Ablation study}

\paragraph{Effects of components.} 
Table~\ref{tab:abl_main} demonstrates the effectiveness of each component in our method. The baseline is shown in ID-1. While the pipeline is similar to previous methods in the Cartesian coordinate system, we input and output in the Polar coordinate system, and only the agent position and lane position are encoded. In ID-2, the agent and lane change embedding is incorporated. By integrating changes in adjacent lane points and changes in agent position (i.e., velocity and acceleration), we observe a notable performance improvement. In ID-3, we add Polar relationship refinement modules using the vanilla Transformer. These modules aim to capture the relationship between the predicted trajectories and the scene context features. To effectively improve prediction accuracy, they need to be combined with the Relative Embedding Transformer. As a result, the performance is only slightly better than that of ID-2, which uses the vanilla Transformer.
Finally, in ID-4, we use the Relative Embedding Transformer to replace the vanilla Transformer. After integrating all these techniques, our model achieves outstanding performance.

\paragraph{Effects of losses.} 
Table~\ref{tab:abl_loss} demonstrates the effectiveness of different loss calculation strategies. We observe that computing the loss solely in either the Polar coordinate system or the Cartesian coordinate system leads to moderate performance. Notably, as shown in the third row, calculating the loss in both coordinate systems simultaneously results in a significant performance improvement. This indicates that leveraging both coordinate systems during training enables the model to make more accurate trajectory predictions.


\begin{figure}[t!]
\centering
\includegraphics[width=0.47\textwidth]{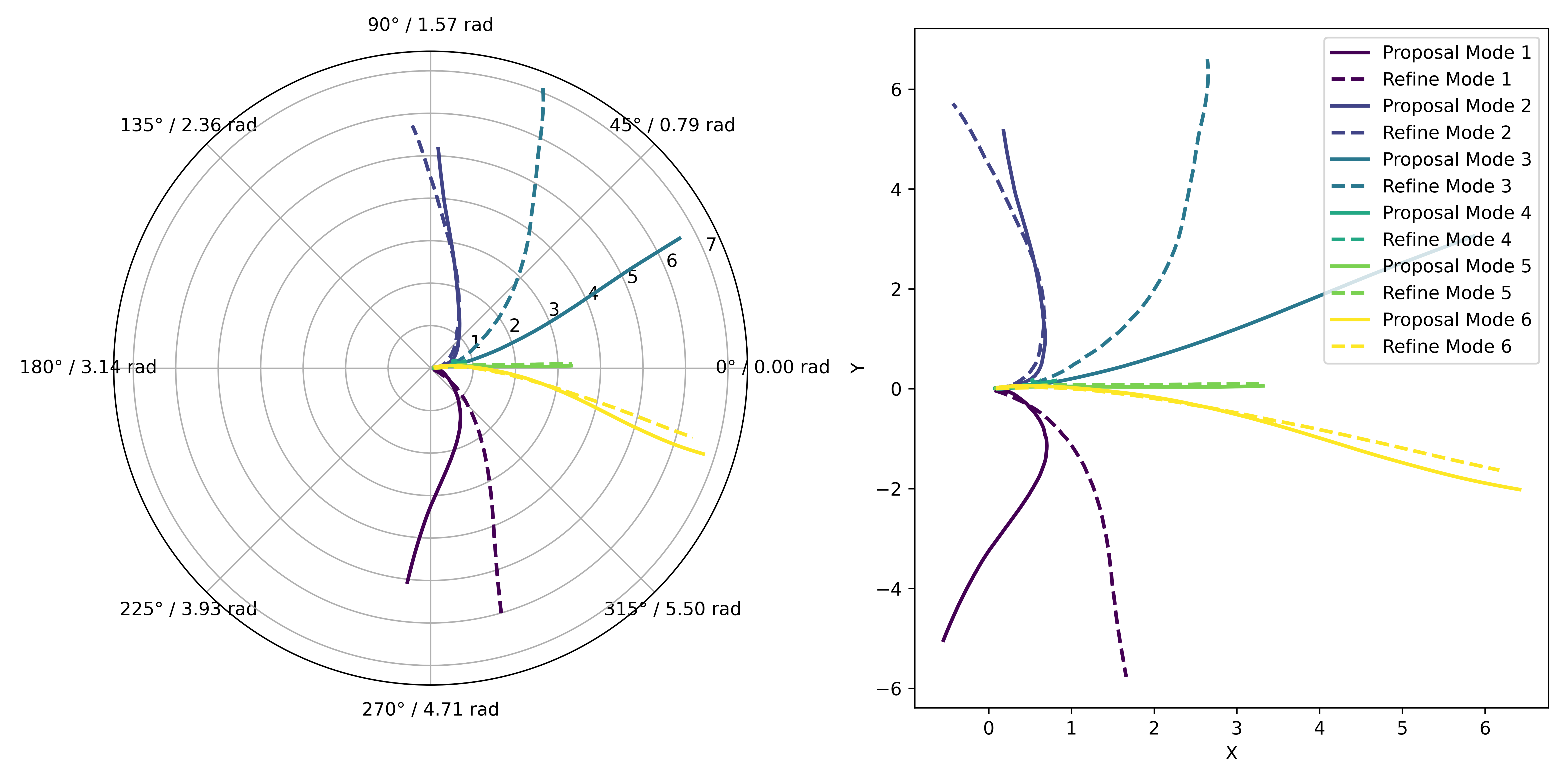}
\caption{
Predicted multi-modal trajectories in the Polar representation (left), and their conversion to Cartesian representation (right). The solid line represents the future trajectory of the proposal, while the dashed line represents the future trajectory after refinement.
}
\label{fig:predict}
\end{figure}

\paragraph{Polar $v.s.$ Cartesian.} 
To highlight the advantages of Polar representation, we conduct an ablation study comparing the performance of Cartesian and Polar representations. 
As shown in Table~\ref{tab:carpol}, the first row represents the original version of the Cartesian representation, where all model components are kept consistent, varying only the input and output in the Cartesian coordinate system using $(x, y)$. To ensure a fair comparison, we still calculate the relative relationships between traffic elements using $(\Delta r, \Delta \theta)$. The results indicate that the Cartesian representation is not naturally suited for extracting relative relationships, such as distances and angles between elements. This requires complex mathematical computations, leading to inference speeds nearly twice as slow as those achieved with Polar coordinates.
In the second row, we modify the architecture for Cartesian-based input, using $(\Delta x, \Delta y)$ to calculate the relative relationships between traffic elements, better aligning with the Cartesian coordinate system, while keeping the other components consistent. The results show that, although faster than the first row, this approach still falls short because the Cartesian representation does not naturally capture the influence of traffic elements based on distance and direction, making it slightly less effective than the first row.

The comparison of results between the first, second, and third rows shows that trajectory prediction and planning in Polar representation better aligns with these relative relationships, leading to superior performance in the third row compared to the Cartesian representation in the first and second rows. This highlights the advantage of our approach.

\subsection{Efficiency analysis and qualitative results}
\paragraph{Efficiency analysis.}
Optimizing the trade-offs between performance, inference speed, and model size is crucial for deployment. We compare our model with recent state-of-the-art models on the Argoverse 2 benchmark. For model size, our model is 4.4M, whereas QCNet~\cite{qcnet} is 7.7M, and SmartRefine~\cite{smartrefine} is 8.0M. Despite its smaller size, our model achieves superior performance compared to these recent high-performance methods. For inference speed, measured on an NVIDIA GeForce RTX 3090 GPU with a batch size of 1, the average inference time of \netName{} is 48 ms, significantly faster than QCNet's 88 ms.


\begin{figure}[t!]
\centering
\includegraphics[width=0.47\textwidth]{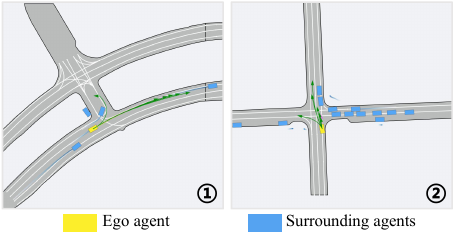}
\caption{
Qualitative results \textit{on the Argoverse 2 validation set}. The predicted trajectories are shown in green, and the ground truth trajectory is shown in pink.
}
\label{fig:vis}
\end{figure}

\paragraph{Qualitative results.}
We present qualitative results to highlight the effectiveness of our model. As shown in Figure~\ref{fig:predict}, we provide an example of predicted multi-modal trajectories in the Polar coordinate system and their conversion to the Cartesian coordinate system. The solid line and the dashed line mean the proposal trajectories and the refined trajectories, respectively.
As shown, predicting multi-modal trajectories involves forecasting trajectory point coordinates, which can result in large variations. However, in the Polar representation, both $r$ and $\theta$ exhibit relatively small changes, making the predictions more stable and accurate compared to the larger variations in $(x,y)$ coordinates in the Cartesian coordinate system.
We can also observe the significant optimization of the trajectories after refinement.
As shown in Figure~\ref{fig:vis}, we also present the trajectory prediction results of our model. The design of our framework enables accurate forecasting of driving behavior. It not only predicts future trajectories precisely but also provides multi-modal outputs, capturing diverse behaviors such as turning, following, and lane-changing for overtaking.
Additional qualitative results and failure cases are provided in the appendix.

\section{Conclusion}
We propose a framework, \netName{}, that advances trajectory prediction and planning by leveraging the Polar coordinate system. Compared to traditional Cartesian-based approaches, our method more effectively captures both the movement of traffic elements and their relative spatial relationships. By integrating Polar scene context encoding and Polar relationship refinement, both utilizing the Relative Embedding Transformer, \netName{} enables precise modeling of interactions between traffic elements. Experiments on Argoverse 2 and nuPlan validate that \netName{} achieves state-of-the-art performance in trajectory prediction and planning.

\paragraph{Limitations and future work.} 
The conversion between Cartesian and Polar coordinates introduces computational overhead, which may affect efficiency. In future work, we aim to explore Polar representations in end-to-end autonomous driving.

\newpage

\section*{Appendix}

\section{Discussions}

\paragraph{Discussion 1:} 
\textbf{Comparison with Cartesian-based methods.} 

As shown in Table~\ref{tab:abl_re_compare}, we provide a comparison between our \netName{} and several representative Cartesian-based methods to highlight the key differences.
In terms of input and output formats, only our \netName{} adopts the Polar representation, whereas MTR~\cite{mtr}, QCNet~\cite{qcnet}, and HPNet~\cite{hpnet} rely on Cartesian coordinates for both input and output.
Regarding the attention mechanism, MTR applies local attention focused on elements near the ego agent, while QCNet and HPNet employ standard (vanilla) attention. In contrast, our \netName{} leverages a Relative Embedding Transformer specifically designed to model relative relationships, fully exploiting the strengths of the Polar representation.
As for the use of relative features, MTR does not utilize them, QCNet and HPNet incorporate relative features in Cartesian form, whereas our \netName{} uses Polar-form relative features expressed as $\Delta r$ and $\Delta \theta$.

This comparison highlights the fundamental differences between our approach and prior Cartesian-based methods.
By leveraging relative relationships and representations in Polar coordinates, our \netName{} introduces a novel modeling paradigm that better captures angular and radial variations in motion, offering a more structured formulation for trajectory prediction and planning. This reflects a core contribution of our work.


\begin{table} [ht!]

\centering
\setlength{\tabcolsep}{1.4mm}
{
\begin{tabular}{l|ccc}
\toprule[1.5pt]
Method & In\&Out & Attention & Relative Feature \\[1.5pt]\hline\noalign{\vskip 2pt}
MTR         & Cartesian & Local Attn. & w/o \\
QCNet       & Cartesian & Vanilla & Cartesian Rel. Feat. \\
HPNet       & Cartesian & Vanilla & Cartesian Rel. Feat. \\
\netName    & Polar & Relative Attn. & $\Delta r$\&$\Delta \theta$ \\
\bottomrule[1.5pt]
\end{tabular}
}
\caption{Comparison of components with related work.}
\label{tab:abl_re_compare}
\vspace{-0.5em}

\end{table}

\paragraph{Discussion 2:} 
\textbf{The use of keypoints in Relative Embedding Transformer.}

As shown in Figure 3 of the main paper, keypoints are encoded to obtain the relative distance and angle embeddings. This operation explicitly incorporates the relative positions among the query, key, and value during attention-based interactions.
In the Polar Scene Context Encoding (Figure 2 of the main paper), we use the centerpoints of agents and lanes as keypoints. The centerpoints of agents represent their current positions, while the centerpoints of lanes correspond to the geometric centers of lane instances.
In the Polar Relationship Refinement (Figure 2 of the main paper), the endpoints of the proposal future trajectories are used as keypoints.

These keypoints represent the most critical spatial anchors for their respective elements, allowing the model to more effectively embed relative positions and capture meaningful spatial relationships among different entities.

\paragraph{Discussion 3:} 
\textbf{Loss calculation in both Cartesian and Polar coordinates.}

We compute the loss in both Cartesian and Polar coordinates. The Cartesian representation provides a simple formulation that facilitates effective and stable convergence during training, while the Polar representation excels at modeling relative relationships. By combining the strengths of both coordinate systems, this dual-loss strategy leads to improved overall performance. The effectiveness of this design is demonstrated in Table 5 of the main paper.

\paragraph{Discussion 4:} 
\textbf{Limitations and future work.}

In addition to the discussion of limitations and future work presented in the conclusion section of the main paper, we provide a more in-depth analysis here to better inform future research directions. Our current framework involves converting inputs and outputs between Cartesian and Polar coordinates, which may introduce additional computational overhead. A promising future direction is to explore the entire autonomous driving pipeline within the Polar coordinate system. This would allow full exploitation of the advantages of Polar representation while avoiding the complexity and inefficiency associated with repeated coordinate transformations. As discussed in the related work section of the main paper, several existing approaches have already adopted Polar representation in the perception module. Therefore, developing a fully end-to-end autonomous driving framework entirely based on Polar representation appears to be a promising research avenue.

\section{Experiment settings}

\subsection{Evaluation metrics}
\paragraph{Trajectory prediction.}
We assess our models using established metrics, including the minimum Average Displacement Error ($minADE_{k}$), minimum Final Displacement Error ($minFDE_{k}$), Miss Rate ($MR_{k}$), and Brier minimum Final Displacement Error ($b$-$minFDE_{k}$). The $minADE_{k}$ metric computes the $L_{2}$ distance between the actual trajectory and the best $K$ predicted trajectories, averaged across all future time steps. The $minFDE_{k}$ metric quantifies the difference between the endpoints of the predicted trajectories and the ground truth. The $MR_{k}$ metric indicates the percentage of scenarios where $minFDE_{k}$ exceeds 2 meters. To offer a more detailed analysis of uncertainty, $b$-$minFDE_{k}$ incorporates $(1-\pi)^2$ into the final displacement error, where $\pi$ represents the probability score given by the model to the best-predicted trajectory. Following the evaluation metrics of the official leaderboard, we set $K$ to 1 and 6 for the Argoverse 2~\cite{argoverse2} and nuPlan~\cite{nuplan} datasets.

\paragraph{Trajectory planning.}
We primarily use the open-loop score (OLS), non-reactive closed-loop score (NR-CLS), and reactive closed-loop score (R-CLS), following the official implementation~\cite{nuplan} and previous works~\cite{PDM,plantf,diffusionplanner,betop}. These scores combine a set of multiplier metrics and weighted average metrics. Each scenario is assigned a score, and the final open-loop score (OLS) or closed-loop score (NR-CLS) is the average of all scenario scores. All metrics fall within the interval $[0, 1]$.

\subsection{Model ensembling}
In our method, we adopt model ensembling—a widely used technique to enhance prediction accuracy. Specifically, we train seven sub-models with different random seeds and training epochs, resulting in 42 predicted future trajectories per agent. To consolidate these predictions, we apply k-means clustering with six cluster centers and compute the average trajectory within each cluster to obtain the final outputs. The results, with and without ensembling, are reported \textit{on the Argoverse 2 test set} in Table 1 of the main paper.


\begin{table} [ht!] 

\centering
{\begin{tabular}{l|c}
\toprule[1.5pt]
$\textbf{Name}$ & $\textbf{Number}$\\
[1.5pt]\hline\noalign{\vskip 2pt}

agent encoding Mamba & 3 \\
encoding Relative Embedding Transformer & 3 \\
decoding vanilla Transformer & 2 \\
Polar Relationship Refinement & 2 \\
refining Relative Embedding Transformer & 2 \\

\bottomrule[1.5pt]
\end{tabular}}
\caption{The number of layers in each component.}

\label{tab:layer}
\end{table}

\subsection{More implementation details}
We present the number of layers in each component in Table~\ref{tab:layer}. Following the implementation details section in the main paper and prior works~\cite{DeMo,plantf,Forecast-mae}, we provide additional details below:

\begin{itemize}
    \item Dropout: 0.2
    \item Activation: GELU
    \item Normalization: LayerNorm
    \item Hidden dimension: 128
    \item Training time: About 20 hours
\end{itemize}


\begin{table} [ht!]

\centering
{
\begin{tabular}{c|ccc}
\toprule[1.5pt]
Number & $\textit{minFDE}_{6}$ & $\textit{minADE}_{6}$ & $\textit{MR}_{6}$\\[1.5pt]\hline\noalign{\vskip 2pt}
0     & 1.33 & 0.69 & 0.17\\
1     & 1.25 & 0.65 & 0.15\\
\rowcolor{gray!20}
2     & 1.21 & 0.63 & 0.14\\
3     & 1.20 & 0.63 & 0.14\\
\bottomrule[1.5pt]
\end{tabular}
}
\caption{Ablation study on the number of the Polar Relationship Refinement modules.}
\label{tab:abl_num}

\end{table}


\begin{table*} [ht!]

\centering
{
\begin{tabular}{l|l|cc}
\toprule[1.5pt]
\textbf{Type} & \textbf{Method} & NR-CLS $\uparrow$ & R-CLS $\uparrow$ \\
[1.5pt]\hline\noalign{\vskip 2pt}

\multirow{2}{*}{Rule} & IDM~\cite{IDM} & 0.77 & 0.76 \\
                      & PDM-Closed~\cite{PDM} & 0.93 & 0.92 \\
\midrule
\multirow{2}{*}{Hybrid} & GameFormer~\cite{gameformer} & 0.83 & 0.84 \\
                        & PDM-Hybrid~\cite{PDM} & 0.93 & 0.92 \\
\midrule
\multirow{7}{*}{Learning} & UrbanDriver~\cite{UrbanDriver} & 0.53 & 0.50 \\
                          & PDM-Open~\cite{PDM} & 0.50 & 0.54 \\   
                          & PlanCNN~\cite{plant} & 0.73 & 0.72 \\
                          & GC-PGP~\cite{GCPGP} & 0.59 & 0.55 \\
                          & PlanTF~\cite{plantf} & 0.85 & 0.77 \\
                          & BeTopNet~\cite{betop} & 0.88 & \bf 0.84 \\
                          & PLUTO~\cite{pluto} & \underline{0.89} & 0.80 \\
                          & DiffusionPlanner~\cite{diffusionplanner} & \underline{0.89} & 0.82 \\
                          \rowcolor{gray!20}
                          & \bf\netName~(Ours) & \bf 0.90 & \underline{0.83} \\
\bottomrule[1.5pt]
\end{tabular}
}
\caption{Performance comparison of closed-loop trajectory planning \textit{on the nuPlan dataset under the Val 14 benchmark}.}
\label{tab:nuplan_val14}

\end{table*}


\begin{table*} [ht!]

\centering
{\begin{tabular}{c|l|c}
\toprule[1.5pt]
$\textbf{Rank}$ & $\textbf{Method}$ & $\textit{b-minFDE}_{6}$ $\downarrow$ \\
[1.5pt]\hline\noalign{\vskip 2pt}

1 & LOF~\cite{LOF} & 1.63 \\
2 & IMR~* & 1.63 \\
3 & SEPT++~* & 1.65 \\
4 & NDPNet~* & 1.71 \\
\rowcolor{gray!20}
5 &~\bf\netName~(Ours) & 1.71 \\
6 & DeMo~\cite{DeMo} & 1.73 \\
7 & SEPT~\cite{sept} & 1.74 \\
8 & DeMo\_PlusPlus~\cite{demo++} & 1.74 \\
9 & X-MotionFormer~* & 1.74 \\
10 & XPredFormer~* & 1.76 \\
11 & DyMap~\cite{DyMap} & 1.78 \\

\bottomrule[1.5pt]
\end{tabular}}
\caption{Argoverse 2 leaderboard at the time of the paper submission. Unreleased works are marked with the symbol ``*". (1) The \#1 method, LOF, is an unpublished work released on arXiv on June 20, 2024. (2) The \#5 method is Our~\netName. (3) The \#7 method, SEPT, is a self-supervised method that utilizes all sets (including the test set) for pretraining and is orthogonal to ours. Our method outperforms all published works on the Argoverse 2 leaderboard (single agent track) at the time of the paper submission (August 2025).} 
\label{tab:av2_leader}
\end{table*}

\section{Experiments}

\subsection{Trajectory planning performance on the nuPlan dataset under the Val 14 benchmark}
To thoroughly evaluate the trajectory planning capabilities of our \netName{}, we present results on the widely used Val 14 benchmark of the nuPlan dataset, as shown in Table~\ref{tab:nuplan_val14}. Our \netName{} achieves performance on par with the current state-of-the-art methods BeTopNet~\cite{betop}, PLUTO~\cite{pluto}, and DiffusionPlanner~\cite{diffusionplanner}, while achieving the best result on the NR-CLS metric.

\subsection{Performance on the Argoverse 2 leaderboard}
As shown in Table~\ref{tab:av2_leader}, we compare our work with others on the Argoverse 2 leaderboard. Our \netName{} outperforms all published works on the Argoverse 2 leaderboard at the time of paper submission (August 2025).

\subsection{Ablation study}
\paragraph{Effects of the depth of refinement modules.} 
Table~\ref{tab:abl_num} shows the number of Polar Relationship Refinement modules. The first row presents the results without the refinement modules. We observe that a depth of two achieves an optimal balance between efficiency and performance.

\section{Qualitative results}

\begin{figure*}[ht!]
\centering
\includegraphics[width=0.68\textwidth]{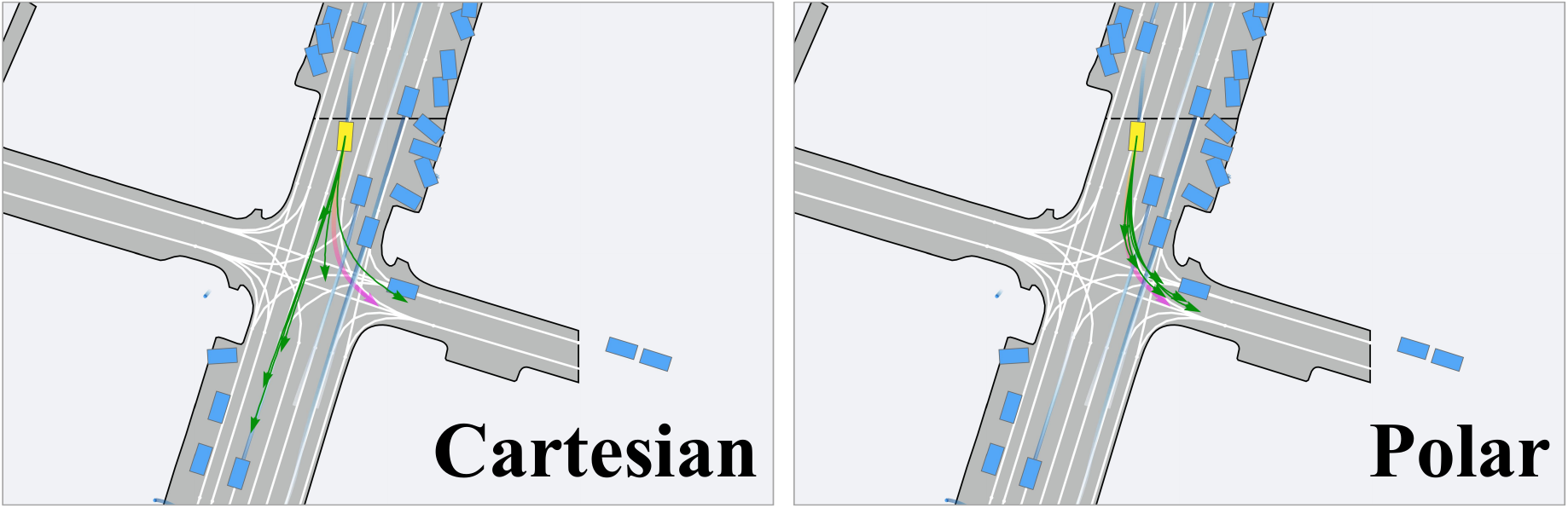}
\caption{
Qualitative comparisons between our \netName{} and the Cartesian baseline \textit{on the Argoverse 2 dataset}.
}
\label{fig:rebuttal}
\end{figure*}

As shown in Figure~\ref{fig:rebuttal}, we present a qualitative comparison between our \netName{} and the Cartesian baseline (introduced in the ablation study of the main paper) to demonstrate the effectiveness of the proposed Polar representation. Leveraging the advantages of Polar coordinates, our \netName{} accurately captures turning intentions by modeling the variations in historical motion, enabling precise prediction of curved trajectories. In contrast, the Cartesian baseline fails to capture such variations and thus produces inaccurate predictions in turning scenarios.

\begin{figure*}[ht!]
\centering
\includegraphics[width=0.68\textwidth]{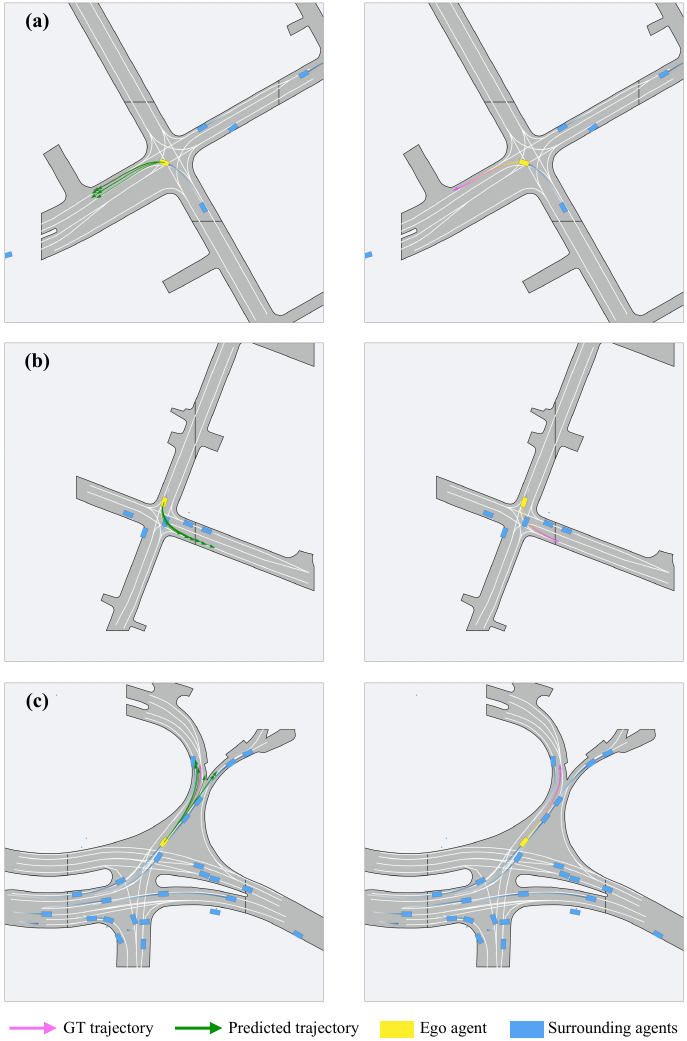}
\caption{
Qualitative results \textit{on the Argoverse 2 validation set}.
}
\label{fig:vissupp}
\end{figure*}

In addition to the qualitative results presented in the main paper, we provide further examples of our \netName{}’s predictions on the Argoverse 2 validation set in Figure~\ref{fig:vissupp}. The left column shows the predicted trajectories, while the right column presents the corresponding ground truth.

\section{Failure cases}
\begin{figure*}[ht!]
\centering
\includegraphics[width=0.68\textwidth]{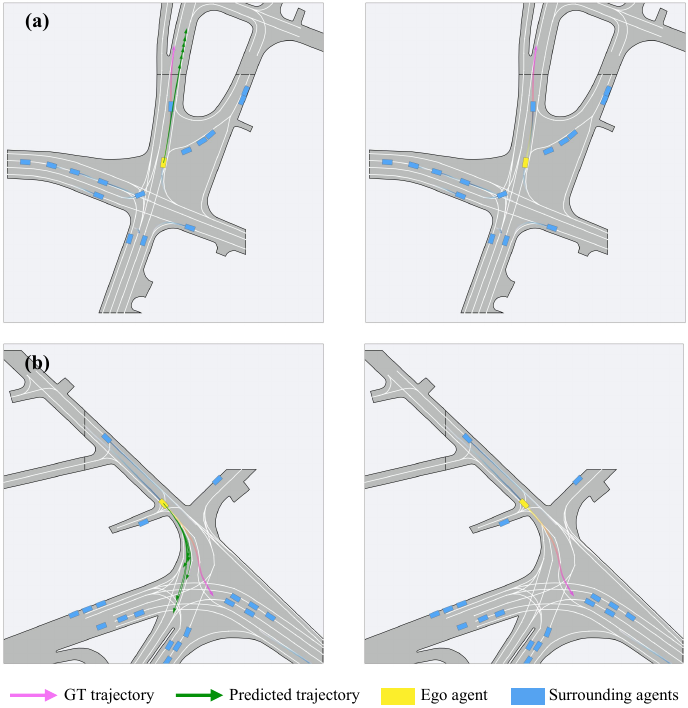}
\caption{
Failure cases \textit{on the Argoverse 2 dataset}.
}
\label{fig:fail}
\end{figure*}

Although our \netName{} has demonstrated strong performance on trajectory prediction and planning benchmarks, certain failure cases still occur, as illustrated in Figure~\ref{fig:fail} using examples from the Argoverse 2 dataset. In Case (a), the model fails to anticipate a lane change, likely due to the absence of explicit cues about the driver’s intention. In Case (b), the model struggles to navigate through a highly complex intersection with multiple potential paths, where ambiguous scene semantics and overlapping road geometries may lead to uncertainty in decision-making.

These failure cases highlight two key limitations: (1) the lack of explicit modeling of driver intent, and (2) the difficulty in reasoning over intricate and cluttered road structures. To address these challenges, future work could explore integrating intent prediction modules or high-level goal inference mechanisms, as well as incorporating richer map representations or hierarchical planning strategies to better handle structural complexity.

\section{Notations}
As shown in Table~\ref{tab:notation}, we provide a lookup table of the notations used in the paper.


\begin{table} [ht!] 

\centering
{\begin{tabular}{c|l}
\toprule[1.5pt]
$\textbf{Notation}$ & $\textbf{Description}$\\
[1.5pt]\hline\noalign{\vskip 2pt}

$M$     &  lanes \\
$dM$     &  lane change variables \\
$A$     &  agents \\

$A_{\rm f}$     &  predicted future trajectories \\
$P_{\rm f}$     &  predicted probabilities \\
$A_{\rm f_{prop}}$     &  predicted proposal future trajectories \\
$P_{\rm f_{prop}}$     &  predicted proposal probabilities \\
$A_{\rm f_{ref}}$     &  predicted refinement future trajectories \\
$P_{\rm f_{ref}}$     &  predicted refinement probabilities \\

$F_{\rm m}$     &  map (lane) features \\
$F_{\rm dm}$     &  lane change features \\
$F_{\rm a}$     &  agent features \\
$F_{\rm s}$     &  scene context features \\

$Q_{\rm traj}$     &  multi-modal trajectory queries \\
$P_{\rm E}$     &  endpoints of predicted trajectories\\

$N_{\rm m}$     &   the number of lane instances\\
$N_{\rm a}$     &   the number of agents\\
$N_{\rm aoi}$     &   the number of agents of interest\\

$C_{\rm m}$     &    feature channels of lanes\\
$C_{\rm a}$     &    feature channels of agents\\
$C$     &    feature channels of latent features\\

$L$     &    the number of points per lane\\
$T_{\rm h}$     &    historical timestamps of agents\\
$T_{\rm f}$     &    future timestamps of agents\\

$K$     &    the number of modalities\\

\bottomrule[1.5pt]
\end{tabular}}
\caption{Notations used in the paper.}

\label{tab:notation}
\end{table}


\bibliography{aaai2026}

\end{document}